\documentclass[letterpaper, 10 pt, conference]{ieeeconf}  % Comment this line out if you need a4paper
\IEEEoverridecommandlockouts
\usepackage{cite}
\usepackage{amsmath,amssymb,amsfonts}
\usepackage{algorithmic}
\usepackage{graphicx}
\usepackage{textcomp}
\usepackage{xcolor}
\def\BibTeX{{\rm B\kern-.05em{\sc i\kern-.025em b}\kern-.08em
    T\kern-.1667em\lower.7ex\hbox{E}\kern-.125emX}}

\title{Energy-efficient Hybrid Model Predictive Trajectory Planning for Autonomous Electric Vehicles}

\author{Fan Ding$^{1}$, Xuewen Luo$^{1}$, Gaoxuan Li$^{1}$, Hwa Hui Tew$^{1}$, Junn Yong Loo$^{1,*}$, \\ Chor Wai Tong$^{2}$, A.S.M Bakibillah$^{3}$, Ziyuan Zhao$^{4}$, Zhiyu Tao$^{5}$ % <-this % stops a space
\thanks{The work of Junn Yong Loo is supported by the Ministry of Higher Education Malaysia under the Fundamental Grant Scheme (FRGS) G-M010-MOH-000206, and Monash University under the SIT Collaborative Research Seed Grants I-M010-SED-000242. $^{*}$ Corresponding author.}
\thanks{$^{1}$The authors are with the School of Information Technology, Monash University Malaysia.
$^{2}$The author is with the School of Engineering, Monash University Malaysia.
$^{3}$The author is with the Department of Systems and Control Engineering, Tokyo Institute of Technology.
$^{4}$The author is with the Institute for Infocomm Research, A*STAR.
$^{5}$The author is with the National Science Library, Chinese Academy of Sciences.}
}

\begin{document}

\maketitle
\thispagestyle{empty}
\pagestyle{empty}

\begin{abstract}
To tackle the twin challenges of limited battery life and lengthy charging durations in electric vehicles (EVs), this paper introduces an Energy-efficient Hybrid Model Predictive Planner (EHMPP), which employs an energy-saving optimization strategy.  EHMPP focuses on refining the design of the motion planner to be seamlessly integrated with the existing automatic driving algorithms, without additional hardware. It has been validated through simulation experiments on the Prescan, CarSim, and Matlab platforms, demonstrating that it can increase passive recovery energy by 11.74\% and effectively track motor speed and acceleration at optimal power. To sum up, EHMPP not only aids in trajectory planning but also significantly boosts energy efficiency in autonomous EVs.
\end{abstract}

% \begin{IEEEkeywords}
% Electric vehicles, autonomous driving, energy-efficiency, kinetic energy recovery systems (KERS)
% \end{IEEEkeywords}

\section{Introduction}
The study of electric vehicles (EVs) has garnered significant attention from both industry and academia, driven by growing concerns about environmental issues \cite{nimesh2021implication}. Autonomous electric vehicles are presented as a practical solution to these challenges \cite{venkitaraman2022review}. However, a widespread adoption faces hurdles such as limited battery life and prolonged charging times \cite{miri2021electric}. To tackle these challenges, researchers are exploring diverse solutions, including innovations in autonomous driving strategies and optimization of route planning algorithms to maximize energy utilization \cite{qin2023review} \cite{taha2018route} \cite{fan2018baidu} \cite{na2023optimal}. Yet, current studies often overlook the interplay among the kinetic energy recovery systems (KERS), engine efficiency, and external environment. These interactions are distinguishing features of hybrid and electric vehicles, which can additionally be incorporated into vehicle driving strategies to improve energy-efficiency\cite{armenta2023analysis}. In prior energy strategy research, driving strategies to improving energy-efficiency have been widely proposed\cite{lv2023impacts}, but it is difficult for drivers to manually control vehicles according to these proposed strategies. The emerge of autonomous driving provides a satisfying solution to the above problems, effective vehicles controlled by algorithms can be more accurately planned and controlled based on strategies.

The conversion of kinetic energy into electrical energy involves reversing the vehicle's electric engine into functioning as a generator and subsequently charging the battery with the generated current \cite{he2022energy}. This conversion of the kinetic energy losses serves as a viable method for energy recovery, enhancing system performance, improving energy conversion efficiency, and extending the mileage \cite{rakov2020study}\cite{ko2014co}\cite{ko2014development}. In recent years, kinetic recovery has emerged as a key focus of interest among researchers, designers, and manufacturers in the EV industry \cite{pan2021kinetic}. As a portion of the energy that is utilized to propel the electric vehicle is dissipated as braking energy loss during driving, recovering a portion of these losses can enhance the efficiency of energy extraction from the battery and prolong the mileage. An alternative approach to regenerative braking for kinetic energy recovery is the conversion of potential energy losses, which can be implemented by not activating the braking system when the vehicle decelerates. In such instances, the kinetic energy recovery system kicks in to recover as much kinetic energy as possible from the car.
Therefore, KERS in energy recovery strategy is an important factor that has to be considered for improving the overall efficiency of EV operation.

In this paper,we propose an energy-efficient Hybrid Model Predictive Planner (EHMPP), optimized based on the best energy efficiency strategies. This planner takes into account the operational state of the KERS during the planning process for EVs, enhancing their range and energy efficiency. We conducted simulation experiments in Prescan, Cars, and Matlab to validate the effectiveness of our proposed strategy. The results indicate that implementing EHMMP significantly enhances the energy efficiency of autonomous electric vehicles.

Our main contributions to this article are as follows:
\begin{itemize}
\item EHMPP is an EVs planner that operates within the constraints of existing hardware configurations, eliminating the need for supplementary hardware deployment.
\item  Our simulation results demonstrate that the proposed strategy significantly enhances the vehicle's energy efficiency during operation. Specifically, it boosts passive energy recovery by 11.74\% during deceleration phases. Moreover, the strategy optimizes motor operation, ensuring it remains close to its ideal power state throughout acceleration, deceleration, and cruising phases, thereby improving overall energy efficiency.
\item 
EHMPP enhances flexibility by automatically selecting distinct cost functions for different motion states, surpassing traditional methodologies. This approach not only facilitates an adaptive planning but also serves as a valuable reference for deploying additional strategies.
\end{itemize}

\section{Related Work}

Over the past decade, the autonomous vehicle industry has undergone remarkable growth \cite{zhang2023alternative}. One that garnered the most attention is trajectory planning \cite{reda2024path}, which is of great significance for charting an accurate course that aligns with a predefined global path. This procedure typically aligns with a predetermined set of control objectives. Simultaneously, vehicles have begun to widely install KERS \cite{yu2019development}, indicating that a transition towards electric vehicles will be the main direction for energy recovery systems in the foreseeable future.

\subsection{Energy Control Strategy}
Incorporating KERS into trajectory planning has ganined immense traction in research, with two prominent approaches: eco-driving control and energy control strategy.  On one hand, examples of eco-driving control approaches include Connected and Autonomous Vehicles (CAVs) which optimizes speed profiles across various scenarios to conserve energy \cite{yang2020efficient}. In addition, Energy Management Systems (EMS) for plug-in hybrid and hybrid vehicles prioritize minimizing powertrain energy consumption while meeting drive power requirements \cite{armenta2023analysis}. Automotive eco-driving is currently regarded as an effective method in enhancing vehicle energy-efficiency without extensive hardware integration \cite{yang2020efficient}. On the other hand, energy control strategy primarily focuses on planning the most energy-efficient trajectory and speed for the vehicle by leveraging a dynamic model of the EV \cite{min2023trajectory}.  For example, the work \cite{xie2019intelligent} introduced an intelligent energy-saving control strategy for EVs, which implements the motion of the preceding vehicle to distinguish between four distinct scenarios and inform control decisions. 
The work cited in \cite{li2021energy} introduces an integrated energy recovery strategy for regenerative braking systems in intelligent, four-wheel independent drive electric vehicles. This approach spans planning to control, addressing energy recovery across three distinct layers. It advocates for trajectory optimization in electric vehicles using an inverse dynamics model and servo constraints, aiming to enhance vehicle energy efficiency through strategic trajectory planning.

Additionally, the work cited in \cite{liu2024eco} presented an energy control strategy for self-driving electric vehicles navigating intersections with continuous speed limit signals. This strategy enhances vehicle energy efficiency under continuous traffic light conditions by regulating vehicle speed.

To our best knowledge, most trajectory planning primarily focuses on the physical conditions and driving strategy of the car, with few works incorporating KERS into planner.

\subsection{Route Planning Algoirthms}

\subsubsection{Dynamic Planning Method}
Dynamic programming (DP), an optimization technique pioneered by Bellman, offers a powerful framework for solving multi-stage decision problems\cite{bellman1966dynamic}. This approach leverages backward recursion, decomposing the problem into a series of single-stage optimizations, starting from the terminal state and work iteratively back to the initial state. While DP has enjoyed success in optimizing energy management for hybrid vehicles by allocating power among the engine, electric motor, and other sources, its computational demands have limited its application in autonomous driving energy strategies. Here, we use a novel approach that leverages DP to obtain a coarse solution, followed by a quadratic program (QP) for refinement.

\subsubsection{Quadratic programming(QP)}
QP is extensively employed for determining the optimal trajectory for autonomous vehicles\cite{li2020road}, with notable advantages over the commonly used nonlinear programming (NLP) methods in trajectory planning. While NLP approaches are versatile, they come with a high computational burden, which can be substantially minimized by adopting QP\cite{liu2021creating}. In contrast, QP not only has reduced computational demands but also excels in speed and convergence when solving convex problems. In particular, trajectory planning requires real-time processing capabilities for rapid dynamic environmental changes.

\section{Methodology}

Our proposed EHMPP framework applies the optimal energy-efficiency strategy on top of DP and QP methods to enhance energy efficiency in autonomous driving. The EHMPP's optimization algorithm involves optimal planning of speed, acceleration, and trajectory, while considering environmental factors such as air resistance, road surface friction, and road elevation. Furthermore, it incorporates real-time vehicle dynamics, such as engine power output and energy recovery system efficiency, into the decision-making matrix.  By establishing distinct cost functions for different acceleration, deceleration and uniform cruising.

\subsection{Constructing an Automobile dynamics model}\label{AA}
Given the central emphasis on energy-efficiency within this paper, the influence of car steering effects is deliberately excluded from consideration in this study.

To achieve simulation control, here we construct an automobile kinematic model that adheres to the following constraints under free motion:
(1) Input parameters of the kinematic model encompass road slope, air resistance coefficient, ground friction coefficient, and car windward area.
(2) Optimal road surface adhesion coefficients ensure balanced adhesion on both sides, mitigating any potential side-slip phenomena.
(3) Motion heading angle constrains the car's trajectory.
(4) The vehicle is treated as a rigid body rather than a mere mass.
The vehicle kinematics model is depicted in Figure 1.

\begin{figure}[t]
    \centering
\includegraphics[width=0.8\linewidth]{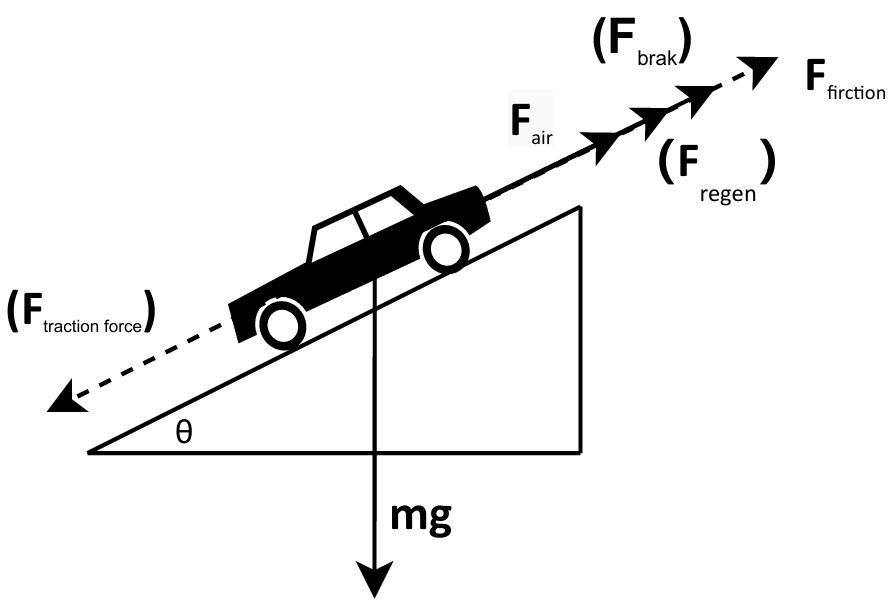}
    \caption{Analyzing Motion Forces on the Car}
    \label{fig:car}
\end{figure}

The air resistance acting on the vehicle is:
\begin{equation}
F_{\text{air}} = \frac{1}{2} \rho v^2 c_d A
\end{equation}
where $\rho$ denotes the air density, $v$ denotes the speed of the car, $C_d$ is the coefficient of air resistance, and $A$ is the windward area of the car.
\begin{equation}
F_{\text{friction}} = \mu mg \cos(\theta)
\end{equation}
The forces induced by slope are: 
\begin{equation}
F_{\text{slope}} = mg\sin(\theta)
\end{equation}
Let \( P_{\text{ENG}} \) and \( P_{\text{REGEN}} \) denote the engine output and maximum power of the energy recovery system, respectively. Thus, the maximum forces from KERS and the motor are \( F_{\text{REGEN}} \) and \( F_{\text{ENG}} \). Since \( F_{\text{REGEN}} \) represents resistance and \( F_{\text{ENG}} \) denotes power, these systems are mutually exclusive, meaning the energy recovery system is inactive during vehicle acceleration.

Consequently, the kinematic relationship during vehicle acceleration can be expressed as follows:
\begin{equation}
ma_{ac} = F_{\text{traction force}} - \frac{1}{2}\rho v^2 c_d A - \mu mg \cos(\theta) \pm mg \sin(\theta)
\end{equation}
The dynamics relationships during deceleration are:
\begin{multline}
ma_{Dec} = -\biggl(\frac{1}{2}\rho v^2 C_d A + \mu mg \cos(\theta) \\
+ F_{\text{REGEN}} + F_{\text{brake}}\biggr)
+ mg \sin(\theta)
\end{multline}

\subsection{optimal energy-efficiency strategy}
Our focus lies in planning of the car's driving and formulating the corresponding cost functions for optimal planning. During the acceleration phase, particular emphasis is placed on achieving energy-saving travel speed and energy-saving acceleration.

It has been shown that maintaining a steady driving behavior results in the lowest energy consumption\cite{galvin2017energy}. This "steady driving behavior" is achieved via accelerating at a low rate to achieve a constant speed, thereby minimizing energy consumption. Furthermore, it has been observed that the relationship between the vehicle's driving speed and energy consumption can be approximated by a quadratic equation.

Therefore, during uniform cruise (where acceleration is zero), the vehicle must adhere to the optimal cruise speed, denoted as \( V_{\text{OPT}} \). This optimal cruise speed, \( V_{\text{OPT}} \), is determined by the vehicle's optimal engine power and external environmental dynamics wich can be expressed as:
\begin{equation}
V_{\text{OPT}} = \frac{F_{\text{Air}} + F_{\text{Friction}} + F_{\text{Slope}}}{P_{\text{Opt}}}
\end{equation}
where \( P_{\text{opt}} \) represents a constant value determined by the electric vehicle's motor.

During acceleration, the EVs should enter the constant speed phase with optimal acceleration without violating constraints.

The constraints for the dynamics model are outlined as follows:
\begin{enumerate}
\item The acceleration of the car must adhere to the limitation of the maximum power output of the motor.
\item Objective constraints govern the braking process of the vehicle.
\item The vehicle operation is mandated by formal road regulations.
\end{enumerate}

According to the studies \cite{galvin2017energy}\cite{wu2015electric}\cite{galvin2017energy}\cite{zhang2020energy}, there is an optimal acceleration with least energy consumption. In this process, the optimal acceleration of the vehicle is modeled via dynamic equation as follows:
\begin{equation}
a_{\text{Acc\_Opt}} = \frac{\frac{P_2}{v} - \left(f_{\text{Air}} + f_{\text{Friction}} + f_{\text{Slope}}\right)}{m}
\end{equation}
where \( P_{\text{2}} \) refers to the optimal output power of the motor or engine in the acceleration phase, which could make acceleration achieve the most energy-efficient.Thus, we identify the acceleration in this state of acceleration as the optimal acceleration.

During deceleration, the EVs decelerates at the optimal deceleration speed fllowing the same constraints.
According to \cite{armenta2023analysis}\cite{demirkale2017investigation}, when the vehicle's kinetic energy recovery system decelerates at a specific speed range, it has the highest kinetic energy recovery efficiency. It is deduced that there is an optimal deceleration in the process of vehicle deceleration, to maximize the efficiency of the kinetic energy recovery system. When this optimal braking condition is met, the vehicle should not apply additional braking force at the brake end. In this process, the optimal vehicle deceleration by dynamic relationship recurrence is as follows:
\begin{equation}
a_{\text{Opt\_Dec}} = \frac{\frac{P_m}{v} + f_{\text{Air}} + f_{\text{Friction}} + f_{\text{Slope}}}{m}
\end{equation}
Where \( P_{\text{m}} \) is the power of the vehicle kinetic energy system with maximum energy-efficiency, which is determined by the configuration of the vehicle kinetic energy recovery system which could make deceleration achieve the most energy-efficient. So we identify the deceleration in this accelerated state as the optimal deceleration.

Our proposed strategy aims to achieve optimal energy efficiency across acceleration, deceleration, and uniform cruising phases. This strategy is designed to optimize the speed and route planning of the autonomous driving system, ensuring alignment with the vehicle’s optimal acceleration and deceleration requirements.

\section{Algorithm implementation}

In our EHMPP, motion planning is categorized into three types based on acceleration states: the acceleration phase with positive acceleration, the constant speed phase with zero drive, and the deceleration phase with negative acceleration.  In particular, the motion/trajectory planning of EHMPP requires a transformation from a Cartesian coordinate system to a Frenet coordinate system, it is necessary to first generate the reference line from the global path which is assumed to be known.

Generally, global paths are not suitable for direct use as reference lines due to their lack of smoothness and excessive length. Therefore, we use the projection point of the car on the global path as the starting point and select an appropriate distance before and after this projection point to smooth and generate the reference line. After obtaining the smooth reference lines, we construct a Frenet coordinate system centered on the car's projection point on the reference line, as illustrated in Figure 2.

\begin{figure}[t]
    \centering
    \includegraphics[width=0.7\linewidth]{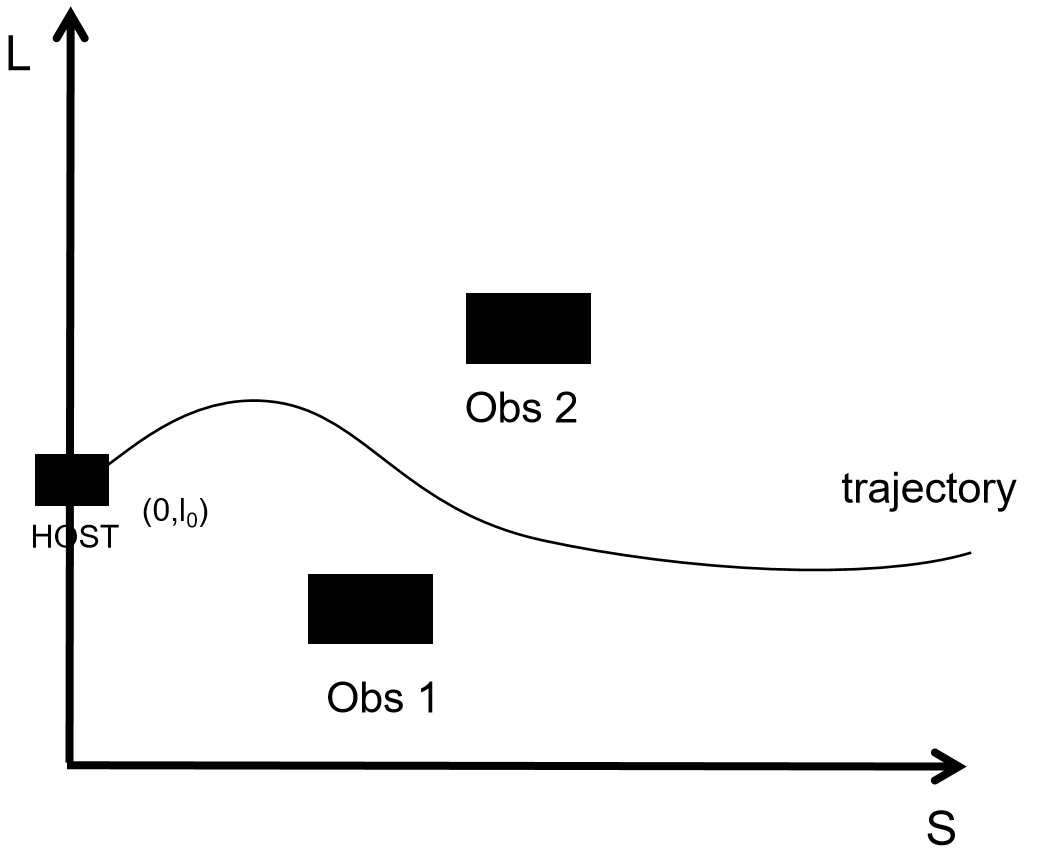}
    \caption{SL of  Frenet coordinate system}
    \label{fig:car}
\end{figure}

At the same time, we project the obstacle onto the established coordinate system. To find the optimal solution within this coordinate state space, directly employing DP or quadratic QP presents several challenges. This is because a direct solution of DP  results in a non-convex optimization problem that is intractable. Therefore, after discretizing the state space, we utilize DP for a rapid search to identify the region where the optimal solution is most likely to exist. So QP can obtain a solution space with monotonicity, called a convex space. Once a rough trajectory is found through DP, QP can be iteratively applied to achieve a refined trajectory.

\subsection{Dynamic programming of trajectories}\label{AA}

We implement a uniform random sampling within a discrete space and subsequently link the sampling points using a quintic polynomial (illustrated in Figure 3). Upon establishing suitable boundary conditions, the trajectory is formulated through the connection of these points with a quintic polynomial.

\begin{figure}[t]
    \centering
    \includegraphics[width=0.75\linewidth]{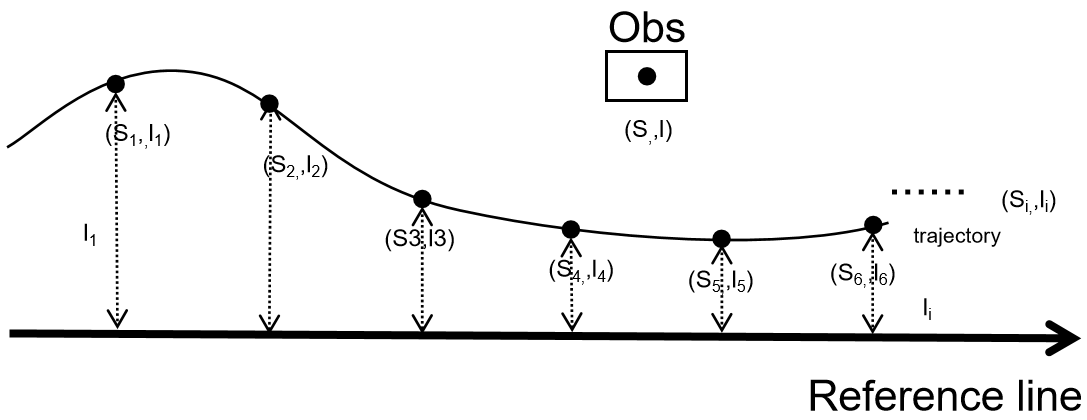}
    \caption{Cost function diagram}
    \label{fig:car}
\end{figure}

The quintic polynomial is expressed as follows:
\begin{equation}
 l = f(s) = a_0 + a_1s + a_2s^2 + a_3s^3 + a_4s^4 + a_5s^5 
\end{equation}

Given a function \( g(s,l) \), the derivatives of the function with respect to the variables \( s \) and \( l \) are denoted as follows:
\begin{itemize}
    \item \( g'(s,l) \) denotes the first derivative of \( g \) with respect to both \( s \) and \( l \).
    \item Higher derivatives follow similarly, e.g., \( g''(s,l) \) for the second derivative.
\end{itemize}

The basic cost function is defined as follows:

\begin{equation}
C_{\text{Obs}}(d)  = 
\begin{cases} 
0 & \text x > d_1 \\
2d + b & \text d_2 < d < d_1 \\
+\infty & \text x < d_2 
\end{cases}
\end{equation}
Where, \( h(d) \) is defined as a segmented function. In the current coordinate system, the variable \( d \) represents the distance between the vehicle and the obstacle. Let the safety interval at the current speed be denoted by the interval \([d_1, d_2]\).
\begin{align}
C_{\text{Sm}}(f) &= w_1 \int (f'(s))^2 \, ds + 
w_2 \int (f''(s))^2 \, ds \nonumber \\
&\quad + w_3 \int (f'''(s))^2 \, ds.
\end{align}

\begin{equation}
C_{\text{Re}}(f) = \int g(s)^2 \, ds
\end{equation}

Equation (11) defines the cost function for the smoothness of the planned trajectory, where \( f'(s_i)^2 \) expresses how similar it is to a straight line. The terms \( f''(s)^2 \, ds \) and \(  f'''(s)^2 \, ds \) quantify the trajectory's curvature and the rate of change of curvature, respectively. Equation (10) details the distance between the vehicle and an obstacle, while Equation (12) describes the distance between the vehicle and the reference line. Additionally, \( C_{\text{sm}} \) denotes the cost associated with trajectory smoothing. The function \( C_{\text{Obs}} \)  calculates the cost related to the distance from obstacles. Define the reference line
 function as $g$$(s)$ and the function \( g(s) \,ds\) \ indicates the cost from the reference line.

In Equation (10), the selection of parameters $d_1$ and $d_2$ critically influences the choice of drag acceleration during the vehicle's deceleration phase and directly impacts the vehicle's energy consumption. The procedure for calculating the optimal values of $d_1$ and $d_2$ for the vehicle is outlined below:
\begin{equation}
F_{\text{Stop\_Max}} = F_{\text{Regen}} + F_{\text{Res}} + F_{\text{Slope}} + F_{\text{Friction}} + F_{\text{Air}}
\end{equation}
Thus,
\begin{equation}
a_{\text{Dec\_Max}} = \frac{F_{\text{Stop\_Max}}}{m}
\end{equation}
\begin{equation}
d_{2} = \frac{v_{f}^{2} - v_{\text{Cur}}^{2}}{2a_{\text{Dec\_Max}}}
\end{equation}
The total cost function of trajectory dynamic programming process is as follows:
\begin{equation}
C_{\text{Total}} = W_{\text{Obs}} C_{\text{Obs}} + W_{\text{Sm}} C_{\text{Sm}} + W_{\text{Re}} C_{\text{Re}}
\end{equation}
In adherence to the energy conservation strategic imperatives of EHMPP, the weighting factor \( W_{\text{Sm}} \) is considerably greater than \( W_{\text{Re}} \). This preference is to ensure the vehicle's trajectory approximates a straight line as closely as possible, thereby mitigating the kinetic energy loss associated with recurrent adjustments of the method disk. After defining the cost function, we use dynamic programming to obtain rough solutions that provide convex space for QP.

\subsection{Quadratic programming of trajectories}\label{AA}

The quadratic programming will search in the convex space opened up by the DP, as shown in Figure 4. QP mainly uses the cost function to find the optimal solution in this convex space, that is, the trajectory is output to the control module.

\begin{figure}[t]
\centering
\includegraphics[width=0.75\linewidth]{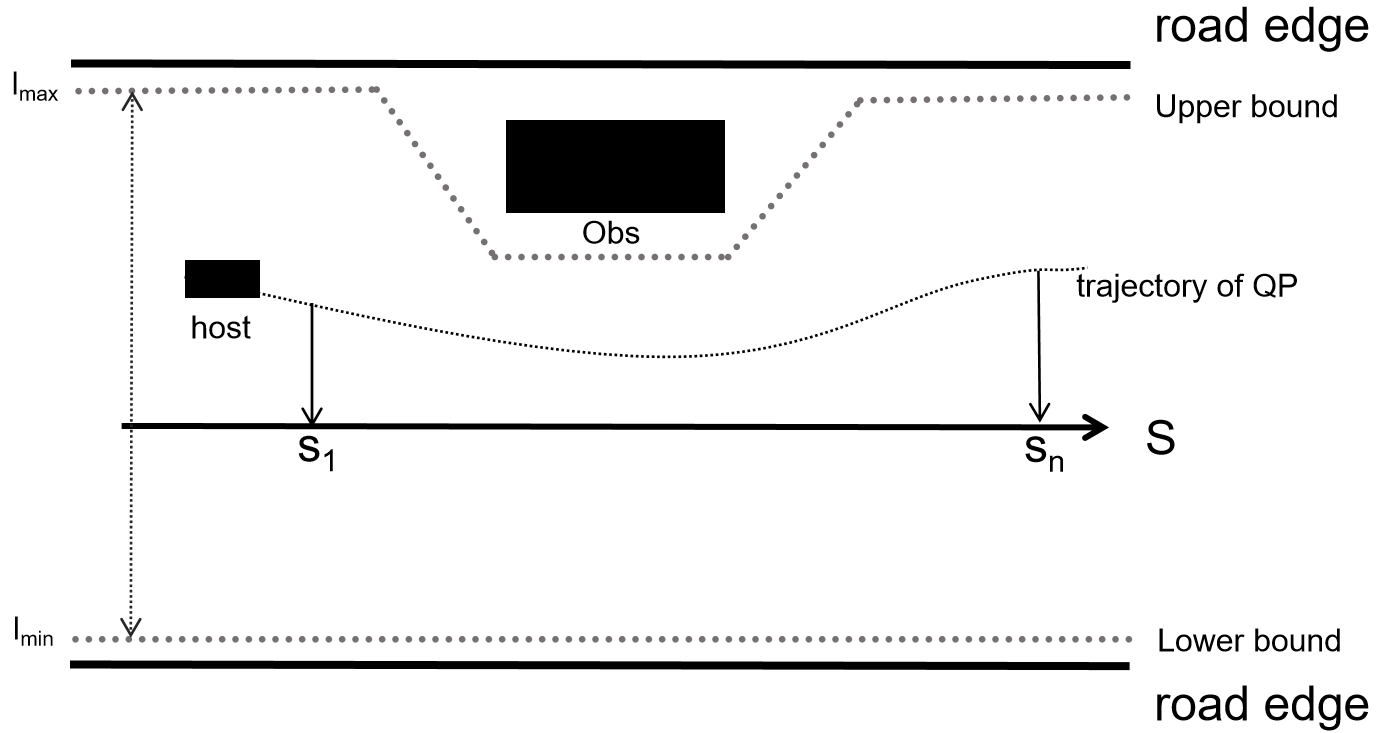}
\caption{QP search space of  Frenet coordinate system}
\label{fig:car}
\end{figure}

In the adopted coordinate framework, the coordinates corresponding to the path point are denoted by \( (s_i, t_i) \), while the subsequent point's coordinates are represented by \( (s_{i+1}, t_{i+1}) \).
QP is utilized to optimize the first and second derivatives of specified points along the trajectory. The trajectory is formulated as a quintic polynomial curve $l$ =  \( f(s) \), with the third derivative maintained as a constant, ensuring that all derivatives of \( f(s) \) of order four and higher are zero between any two consecutive points \( i \) and \( i+1 \). Subject to the continuity constraints on the second derivative of \( f(s) \) and vehicular collision avoidance, the trajectory optimization cost, following a finite-term Taylor expansion at points \( i \) and \( i+1 \).
Upon incorporating the cost near the reference line, as formulated through quadratic programming, is delineated as follows:

\begin{align}
C_{\text{total}}(f) = & \, w_1 \int (f'(s))^2 \, ds + w_2 \int (f''(s))^2 \, ds \notag \\
                      &+ w_3 \int (f'''(s))^2 \, ds
                      + w_4 \int (f(s) - g(s))^2 \, ds.
\end{align}

Where, \( g(s) \) refers to the rough solution trajectory of DP.
The cost function \( (f(s) - g(s))^2 \) quantifies the cost to the reference line. A greater path deviation results in an increased cost function value, thereby imposing a higher deviation penalty. The \( (f'(s))^2 \) represents the cost function for trajectory smoothness. The \( (f''(s))^2 \) is the curvature cost and \( (f'''(s))^2 \)   is the curvature continuity cost.

After completing trajectory planning, we need to carry out velocity planning on this basis.

\subsection{Velocity planning}\label{AA}

Similar to trajectory planning, velocity planning requires a combination of dynamic and quadratic programming for similar reasons.
A new Frenet coordinate system is established with the trajectory as the coordinate axis. In order to simplify the calculation, we regard the obstacle as a particle and construct the ST diagram as follows:

\begin{figure}[t]
    \centering
    \includegraphics[width=0.5\linewidth]{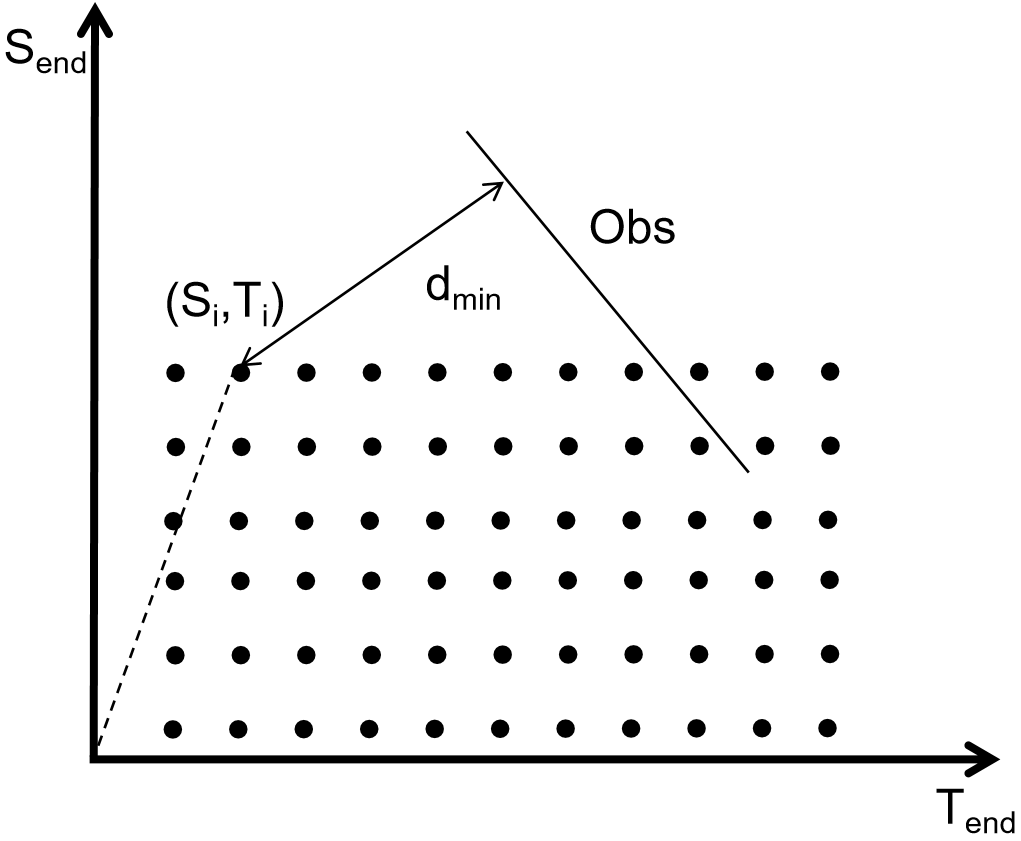}
    \caption{SL diagram of speed planning}
    \label{fig:car}
\end{figure}

In EHMPP, the acceleration stage cost function comprises three components: obstacle, optimal acceleration, and recommended speed costs.
 In the space-time (ST) diagram, the generated trajectory is represented by $S$$(t)$ .Simplified within the ST diagram, the obstacle cost represents the minimal distance from a point to a line segment.To differentiate from the previous stage, we denote the derivative of the function \( f(s) \), representing acceleration, by \( \dot{s}_i \) and so forth.
 Accordingly, the obstacle cost function is derived as follows:
\begin{equation}
C_{\text{obs}} = 
\begin{cases} 
0 & \text{if } |d_{\text{min}}| > d_1 \\
\frac{A}{d_{\text{min}} - d_2} & \text{if } d_2 < |d_{\text{min}}| < d_1 \\
+\infty & \text{if } |d_{\text{min}}| < d_2
\end{cases}
\end{equation}
Where,  \( d_{\text{min}} \) refers to the minimum distance from the obstacle.

For simplicity, the derivatives are approx-imated by the finite difference method. Simultaneously, we discretize the trajectories into discrete points and evaluate the associated cost function. The cost function for reference speed of the point is defined as:
\[
C_{\text{ref\_speed}} = W_{\text{ref\_speed}} (\dot{s}_i - v_{\text{opt}})^2
\]
where \(\dot{s}_i\) represents the actual speed of the vehicle, and \(v_{\text{opt}}\) is the optimal speed.

EHMPP aims to maintain as Optimal acceleration  as possible and accelerate gently to achieve optimal acceleration energy consumption:
\[
C_{\text{acc}} = W_{\text{acc}} (\ddot{s}_i - a_{\text{acc\_opt}})^2 + W_{\text{je}} (\dddot{s}_i)^2
\]
where \(\ddot{s}_i\) is the actual acceleration, and \(\dddot{s}_i\) is the jerk of the vehicle.
Therefore, we define the cost function of speed planning in the acceleration phase as follows:
\begin{equation}
C_{\text{acc\_total}} = W_1 C_{\text{ref\_speed}} + W_2 C_{\text{acc}} + W_3 C_{\text{obs}}
\end{equation}

In the deceleration phase, EHMPP considers not only the minimum braking distance within the path planning phase but also the optimal drag acceleration during deceleration. This approach ensures the kinetic energy recovery system operates at maximal power. The cost function for acceleration in the deceleration phase is defined accordingly:
\[
C_{\text{dec}} = W_{\text{acc}} (\ddot{s}_i - a_{\text{dec\_opt}})^2 + W_{\text{je}} (\dddot{s}_i)^2
\]
At the same time, obstacle cost and optimal speed cost need to be considered. The overall cost function is constructed as follows:
\begin{equation}
C_{\text{dec\_total}} = W_1 C_{\text{ref\_speed}} + W_2 C_{\text{dec}} + W_3 C_{\text{obs}}
\end{equation}

During constant cruising, EHMPP solely considers the recommended speed cost:
\begin{equation}
C_{\text{con\_total}} = W_1 C_{\text{ref\_speed}}  + W_3 C_{\text{obs}}
\end{equation}

Upon calculating the cost for all points, the endpoint for velocity planning is determined by traversing the upper and right boundaries to identify the point with minimal cost. Subsequently, the preliminary solution for the dynamic programming model is derived through a reverse solution process.

After deriving an initial solution, the QP algorithm extends its search within the opened convex space. At this juncture, QP primarily functions to refine the solution, ensuring smoothness and adherence to predefined constraints. The cost function employed by QP at each stage is detailed below:

Acceleration Phase Cost:
\begin{equation}
C_{\text{QP\_acc}} = \sum (W_1 C_{\text{ref\_speed}} + W_2 C_{\text{acc}} + W_3 C_{\text{obs}})
\end{equation}
Deceleration Phase Cost:
\begin{equation}
C_{\text{QP\_dec}} = \sum (W_1 C_{\text{ref\_speed}} + W_2 C_{\text{dec}} + W_3 C_{\text{obs}})
\end{equation}
Constant Cruising Phase Cost:
\begin{equation}
C_{\text{QP\_cru}} = \sum (W_1 C_{\text{ref\_speed}} + W_3 C_{\text{obs}})
\end{equation}
The subsequent processing of the generated trajectory and velocity plans falls outside the scope of this discourse and, as such, will not be elaborated upon.

\section{Experimental data analysis}
In the experimental environment, Matlab, Carsim, and Perscan were used to conduct joint simulations to verify the usability and results of the strategy in the general environment.
The experimental simulation environment covers a series of commands carried out by the car during normal driving, including acceleration, deceleration, uniform speed, lane change, and avoidance. We conduct a comparative analysis of the traditional planner and EHMPP within the same testing environment.
After the above experiments, we obtained the time-changing velocity and acceleration of the car before and after the improvement, as well as the planned trajectory. Since this paper emphasizes the influence of strategy on velocity planning, and velocity is influenced by acceleration, we discuss acceleration based on the conclusion.
According to the formula, the motor output power of the electric vehicle during operation can be calculated as:
\begin{equation}
P = F \cdot V
\end{equation}
F is the total force of the motor acting on the car, and V is the speed of the car. In this experiment, in order to visually see the effect of the experiment and facilitate calculation, we ignore other forces and only consider the motor power, braking force, and kinetic energy recovery force of the car.
The approximate calculation formula of Power (P) is as follows:
\begin{equation}
P \propto |a| \times |V|
\end{equation}

In the sampling time, the power is calculus with the sampling time, which is approximately regarded as the total power of the whole process:
\begin{equation}
E = \int P(t) \, dt
\end{equation}

\begin{figure}[htp]
\centering
\includegraphics[width=0.5\linewidth]{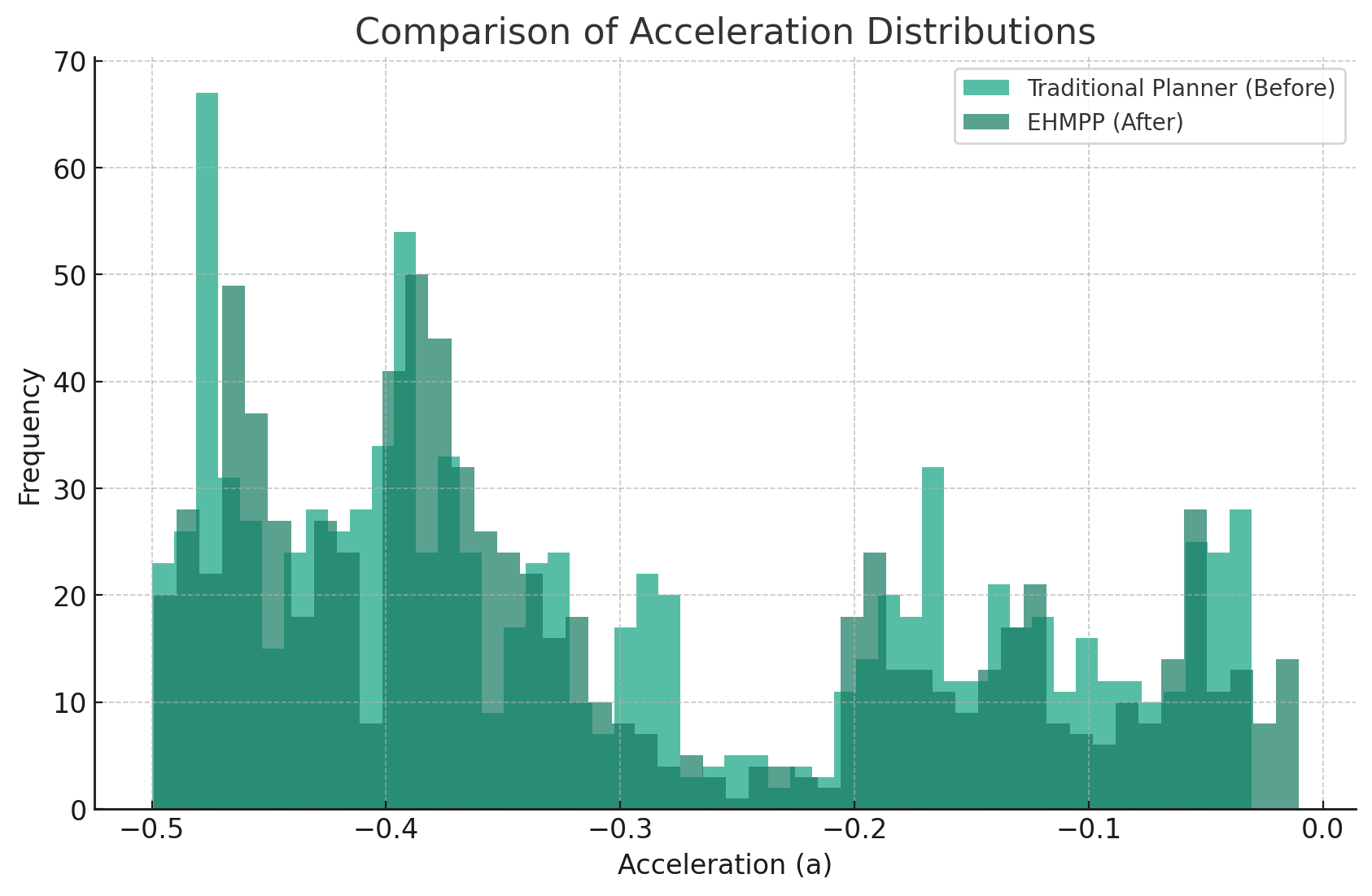}
\caption{Acceleration frequency distribution diagram}
\label{fig:car}
\end{figure}

Because the data collected before and after were in the same simulation situation, other conditions except the strategy were the same (same vehicle, same road condition, same distance traveled, same sampling time, same recommended speed, and cost weight). Therefore, to simplify the calculation, we can discretize the experimental data set:
Deceleration stage:
In the experiment, the kinetic energy recovery system can provide drag acceleration within the range of 0 to -3, beyond which it needs to rely on vehicle braking. At the same time, the kinetic energy recovery system cannot start to work if the drag acceleration is lower than 0.5 in the deceleration stage, as follows:
\begin{enumerate}
\item With all conditions equal, vehicle drag acceleration between 0 and -0.5 is reduced by 11.74\% after deploying the strategy (Figure 6). This indicates that autonomous driving planning, referencing the policy, is biased to decelerate using drag acceleration greater than 0.5, consistent with planning.

\item In the simulation experiment, during the first half of the busy traffic scenario, the vehicle's EHMPP demonstrated performance consistent with that of the traditional planner, maintaining effective obstacle avoidance. However, in the latter half of the day, when traffic conditions were lighter, the EHMPP achieved reduced energy consumption (Figure 7).

\item The indirect optimization of energy consumption by EHMPP allows the vehicle to sustain optimal output power, thereby enhancing overall energy efficiency and reducing energy consumption.
\end{enumerate}

\begin{figure}[htp]
    \centering
    \includegraphics[width=0.75\linewidth]{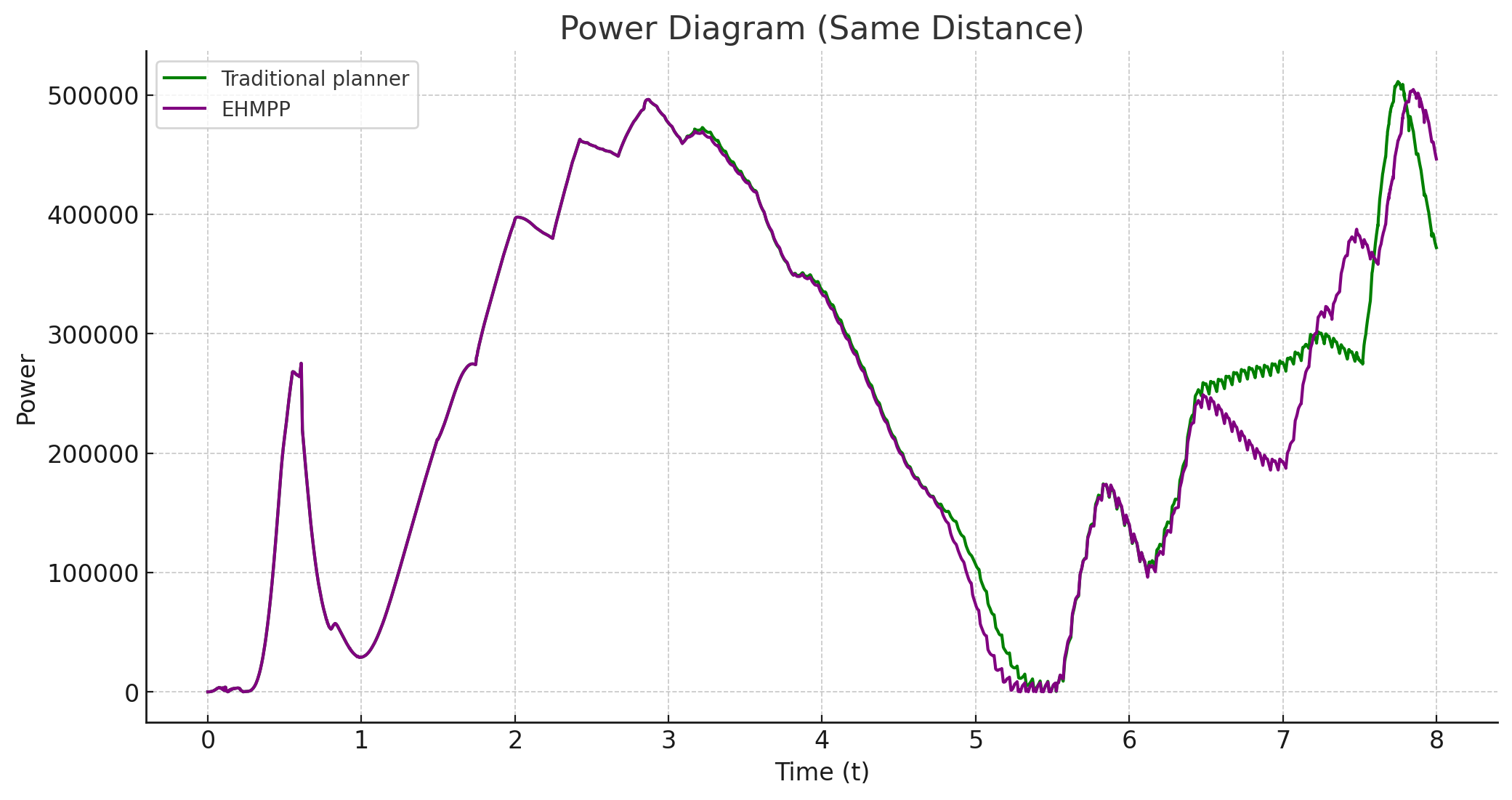}
    \caption{Power diagram (same distance)}
    \label{fig:car}
\end{figure}

In summary, experiments show that EHMPP significantly improves energy efficiency compared to traditional planners.

\section{Conclusions and future works}

This work introduced EHMPP, a hybrid model predictive planner designed to enhance vehicle energy efficiency, address mileage anxiety, and maximize the utilization of vehicle KERS.  The core feature of this planner it considers different vehicle motion states during the planning process and apply distinct optimization equations for each state.  EHMPP has been validated in an experimental environment and demonstrated a substantial energy-saving effect.

EHMPP is capable of handling basic autonomous driving scenarios, including the avoidance of multiple static and dynamic obstacles.  It plans trajectories by predicting the movements of obstacles over a set period.  Future enhancements to EHMPP could include additional modules to accommodate more complex driving scenarios, further optimizing the energy efficiency of EVs.  Additionally, the concept of maintaining optimal power for each device of EVs could be extended to other driving decision planners to optimize their energy consumption.
\bibliographystyle{ieeetr}
\bibliography{ref}

\end{document}